\pdfoutput=1

\documentclass[11pt]{article}

\usepackage[final]{acl}

\usepackage{times}
\usepackage{latexsym}

\usepackage[T1]{fontenc}

\usepackage[utf8]{inputenc}

\usepackage{microtype}

\usepackage{multirow}
\usepackage{tabularx}
\usepackage[switch]{lineno}
\usepackage{algorithm}
\usepackage{algorithmic}
\usepackage{amsmath}

\usepackage{booktabs}
\usepackage{longtable}
\usepackage{multicol}
\usepackage{graphicx}

\title{Aligning Language Models with Offline Learning from Human Feedback}

\author{ Jian Hu \\ NVIDIA \\ \texttt{jianh@nvidia.com} \\\And
  Li Tao \thanks{Equal contribution} \\ NVIDIA \\ \texttt{lit@nvidia.com} \\\And 
  June Yang  \\ NVIDIA \\\texttt{juney@nvidia.com} \\\And 
  Chandler Zhou \\ NVIDIA \\ \texttt{chandlerz@nvidia}
    }

\begin{document}
\maketitle
\begin{abstract}
Learning from human preferences is crucial for language models (LMs) to effectively cater to human needs and societal values. Previous research has made notable progress by leveraging human feedback to follow instructions. However, these approaches rely primarily on online learning techniques like Proximal Policy Optimization (PPO), which have been proven unstable and challenging to tune for language models. Moreover, PPO requires complex distributed system implementation, hindering the efficiency of large-scale distributed training. In this study, we propose an offline learning from human feedback framework to align LMs without interacting with environments. Specifically, we explore filtering alignment (FA), reward-weighted regression (RWR), and conditional alignment (CA) to align language models to human preferences. By employing a loss function similar to supervised fine-tuning, our methods ensure more stable model training than PPO with a simple machine learning system~(MLSys) and much fewer (around 9\%) computing resources. Experimental results demonstrate that conditional alignment outperforms other offline alignment methods and is comparable to PPO. 
\end{abstract}

\section{Introduction}

Recently, advances in language models (LMs) have revolutionized natural language processing, unlocking unprecedented capabilities in text generation. Although these models are powerful, they sometimes produce outputs that diverge from human desirability, like being helpful, not offensive~\cite{bai2022constitutional}, truthful, and fair. An important challenge is to make language models align with human values and preferences. This means adapting them so the generated texts match what a person intends or is seen as helpful, honest, and harmless. Researchers are working on ways to get language models to produce texts that follow human ethics and social norms~\cite{ouyang2022training}. 

One of the most critical elements in achieving this is the use of human feedback, and the most famous approach enabling language models to align with human intent is Reinforcement Learning with Human Feedback (RLHF). RLHF used online learning algorithms~\cite{ouyang2022training}, such as Proximal Policy Optimization (PPO)~\cite{schulman2017proximal}, combined with reward models trained on human feedback to fine-tune language models. However, PPO faces challenges in training stability and difficulty tuning hyper-parameters, such as seeds, Kullback-Leibler (KL) divergence penalty, learning rate, batch size, and other implementation tricks~\cite{islam2017reproducibility,henderson2018did,shengyi2022the37implementation,hu2021rethinking}. Inappropriate hyper-parameters can cause the PPO policy to collapse. Furthermore, implementing the PPO algorithm for language modeling in a large-scale distributed setting is complex, necessitating communication and coordination across multiple modules~\cite{nemorlhf2023}, such as the actor, critic, initialized policy, and reward models shown in Figure~\ref{img:rlhf}. This adds complexity that can hinder the efficiency of the large-scale training system. In addition, the actor and critic modules are involved in both the training and inference tasks for sample generation, which further increases the difficulty of optimizing the system.

In this work, we propose an offline alignment framework without interacting with environments and three offline alignment algorithms: filtering alignment (FA), reward-weighted regression (RWR)~\cite{peters2007reinforcement}, and conditional alignment (CA). For filtering alignment, reward scores are used to filter samples, so that only high-quality samples will be used in the alignment training. For RWR, reward scores will be used to adjust the loss, in which condition high-reward samples have more impact on the parameter updating. For conditional alignment, it introduces a policy based on the cross-entropy method that makes training more stable and efficient. For a concrete implementation, we first train a high-quality reward model using human preference datasets. The reward model is then used to label the finetune samples with rewards. At last, we use the rewards to finetune the language model with the methods mentioned above. By employing a loss function similar to supervised fine-tuning, offline alignment methods can train with a simple machine learning system (MLSys) and much fewer computing resources than PPO, which is suitable for fast alignment training experiments. Our experimental results show that conditional alignment performs better than other offline alignment methods and is comparable to PPO. In summary, our offline alignment framework enables a more efficient and stable alignment of language models to human preferences without complex distributed RL systems.

\section{Related Works}
\subsection{Language Model Alignment with Human Feedback}
ChatGPT~\cite{chatgpt2022,ouyang2022training} trains a large language model (LLM) based on a pre-trained Generative Pre-trained Transformer (GPT)-3.5~\cite{brown2020language} model in 3 steps, supervised fine-tuning, reward model training, and PPO training.

\noindent\textbf{Supervised Fine-tuning (SFT)} The researchers fine-tuned GPT-3 on human demonstrations from their labelers using supervised learning loss in Eq.~\ref{eq:sft}.

    \begin{equation}
    loss(\phi)=-\sum_i \log p_\phi\left(x_{i} \mid p, x_{<i}\right),
    \label{eq:sft}
    \end{equation}

    where $x_{i}$ is the $i_{th}$ token in the sequence, and $p$ is the human instructions and prompts.
    
\noindent\textbf{Reward Model (RM) training} Starting from the SFT model with the final unembedding layer removed, the researchers trained a model to take in a prompt and response and output a scalar reward. Specifically, the loss function for the reward model is,
    
    \begin{equation}
        \begin{aligned}
        &loss(\theta)= \\ &-E_{\left(x, y_w, y_l\right) \sim D}\left[\log \left(\sigma\left(r_\theta\left(x, y_w\right)-r_\theta\left(x, y_l\right)\right)\right)\right],
            \end{aligned}
    \label{eq:rank_loss}
    \end{equation}
    
    where $r_\theta (x, y)$ is the scalar output of the reward model with parameters $\theta$ for prompt $x$ and response $y$, $y_w$ is the preferred response out of the pair of $y_w$ and $y_l$, and $D$ is the dataset of human comparisons.

\noindent\textbf{PPO training} The researchers fine-tuned the SFT model on their bandit environment using PPO. In this environment, a random customer prompt is presented and a response is expected. The environment then produces a reward determined by the reward model and ends the episode, given the prompt-response pair. Additionally, a per-token KL penalty from the SFT model is added at each token to mitigate over-optimization of the reward model. The value function is initialized from the RM.
    
    The bandit environment enables directly optimizing the SFT model for reward from the pre-trained reward model while regularizing against diverging too far from the original SFT through the KL penalty. Initializing the value function from the RM provides a stable starting point for RL fine-tuning. The loss function of PPO-ptx is shown in Eq.~\ref{eq:ppo}.
    
    \begin{equation}
    \begin{aligned}
        & objective(\phi)= \\ &E_{(x, y)  \sim D_{\pi_\phi^{\mathrm{RL}}}} \left[r_\theta(x, y)  - \beta \log \left(\pi_\phi^{\mathrm{RL}}(y \mid x) / \pi^{\mathrm{SFT}}(y \mid x)\right)\right] \\ + 
    & \gamma E_{x \sim D_{\text {pretrain }}}\left[\log \left(\pi_\phi^{\mathrm{RL}}(x)\right)\right],
        \end{aligned}
    \label{eq:ppo}
    \end{equation}

    where $\pi^{\mathrm{RL}}$ is the learned reinforcement learning policy, $\pi^{\mathrm{SFT}}$ is the supervised fine-tuned model, and $\beta$ is the KL reward coefficient that controls the strength of the KL penalty. The $D_{\text {pretrain}}$ is the pretraining distribution and the $\gamma$ is the coefficient that controls the strength of pre-train loss which aim to fix the performance regressions on public NLP datasets. When $\gamma$ is set to $0$, we call this algorithm PPO.

\begin{figure}[htbp]
\centering
\includegraphics[width=1.0\columnwidth]{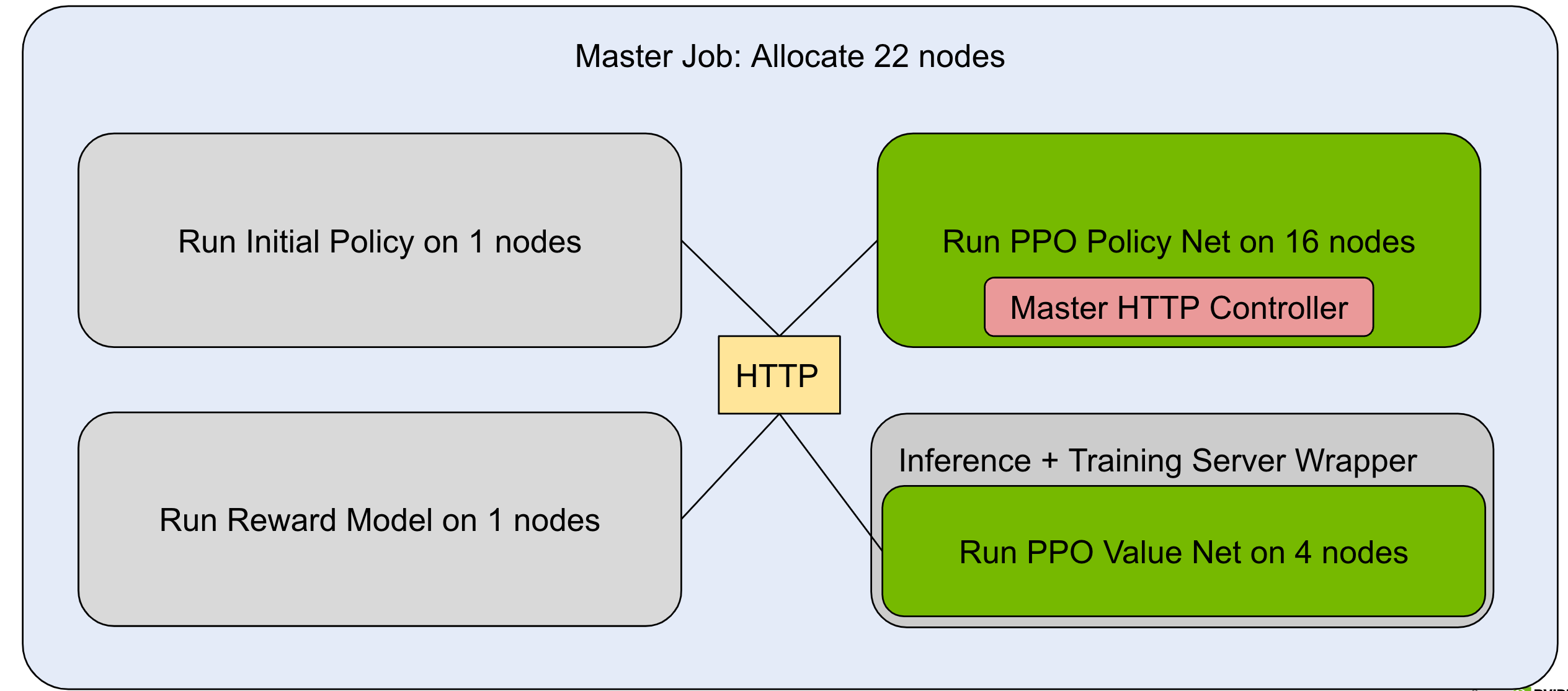} 
\caption{NVIDIA NeMo Aligner~\cite{nemorlhf2023} for language models across 22 nodes - 22B for the actor and policy models and 8B for the reward and value models. Each node is a DGXA100 computer consisting of 8 NVIDIA A100 Graphics Processing Units (GPUs) connected with NVLink~\cite{li2019evaluating}. The gray nodes are used solely for inference during PPO sample generation, while the green nodes handle both sample generation and training. These modules communicate with each other using the HyperText Transfer Protocol (HTTP).}
\label{img:rlhf}
\end{figure}


\subsection{Distributed PPO Training System for Large Models}

There are some open-source RLHF training frameworks such as ColossalChat \footnote{\url{https://chat.colossalai.org/}} and DeepSpeedChat~\cite{yao2023deepspeedchat}, which are developed based on ZeRO~\cite{rajbhandari2020zero}. However, these frameworks are not suitable for training extremely large models, such as 70B models \footnote{Although DeepSpeedChat has demonstrated training of 60B models, their critic and reward model size is only 350 million parameters.}. This is because, with the large scale of all four models in RLHF, it would be challenging to load them into shared GPU memory. For large models, the PPO training system is typically designed with multiple communicable modules to address the issue that a single node cannot load four large models. Such as the NVIDIA NeMo Aligner~\cite{nemorlhf2023} which was developed on NeMo Megatron~\cite{shoeybi2019megatron} consists of four modules - the actor, critic, initialized policy, and reward models. 

As shown in Figure~\ref{img:rlhf}, these four modules need to communicate with each other during training. And during PPO training, the actor module generates samples based on customer prompts and distributes them to the reward, initialized policy, and critic modules. This stage we call the generation stage often involves optimizing inference performance, including key-value (KV) cache~\cite{transformerinference2022}, Flash-Attention~\cite{dao2022flashattention}, tensor layout conversion~\cite{vanholder2016efficient} for the inference engine, etc., which increases the complexity of the whole system. It then gathers the returns, initial log probabilities, and values from these modules to assemble the full PPO training samples in the rollout buffer. Finally, the PPO samples are sent to the actor and critic modules to calculate the PPO loss for training.

While this architecture design allows large models to be trained, it also introduces challenges in improving GPU utilization efficiency and overall training efficiency due to the system's complexity. Optimizing the interactions to reduce communication overhead, such as using asynchronous communication between the Actor module and other modules, and maximizing single-node efficiency, is crucial to enhancing overall performance. However, this complexity impacts both the efficiency of model training and the ease of use of the training framework.

\begin{figure*}[htbp]
\centering
\includegraphics[width=1.7\columnwidth]{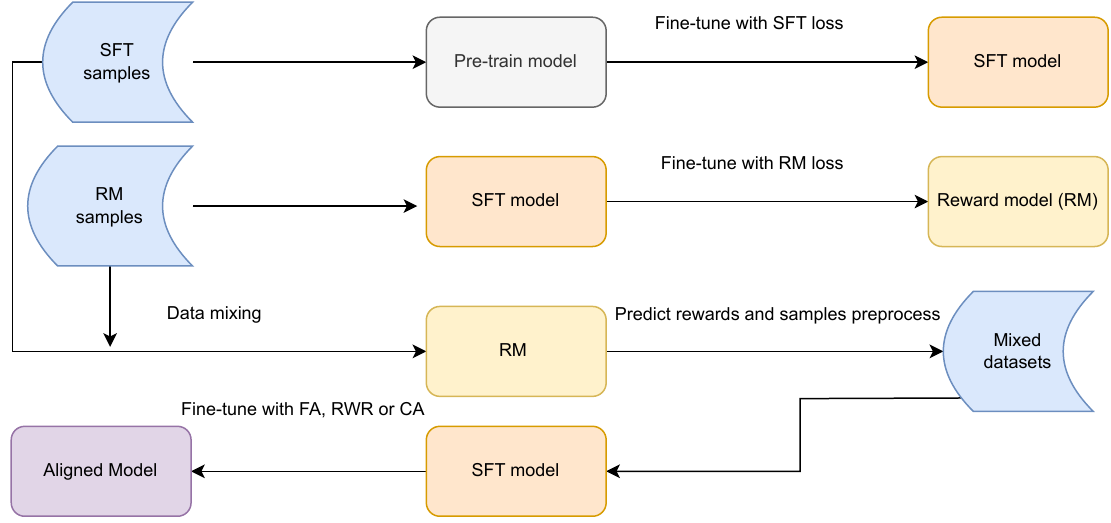} 
\caption{The workflow of the offline alignment framework.}
\label{img:offline_rlhf}
\end{figure*}

\section{Offline Learning from Human Feedback} \label{sec:offlinerlhf}

In this section, we propose an offline alignment framework - that is the language model is fine-tuned on pre-generated samples for the human intent alignment, without the environment interactions in online learning. The training process of offline alignment can be described in the following steps and illustrated in Figure~\ref{img:offline_rlhf}.

\subsection{Offline Alignment Steps}

    \noindent\textbf{1. Supervised Finetune} Similar to ChatGPT, we first fine-tune the pre-trained model on human-labeled instruction data using supervised learning. This allows the model, which we call SFT, to learn the format and intent of human instructions.
    
    \noindent\textbf{2. Reward Model Training} We also train a separate human preference model on available loss functions such as binary loss or ranking loss (Eq. ~\ref{eq:rank_loss}). This preference model learns to predict rewards.
    
    \noindent\textbf{3. Training Samples Labeling} We utilize the reward model to label the collected instruction samples, including both human-labeled samples and pair-wise samples from reward model training. 
    
    \noindent\textbf{4. Alignment Finetune} Finally, we fine-tune the SFT model using offline learning algorithms on the reward-labeled training samples to align the model with human preferences. The instructions and loss functions depend on which offline alignment method we use (i.e., FA, RWR, or CA). 

We also propose three offline alignment algorithms: 1) filtering alignment, 2) reward-weighted regression~\cite{peters2007reinforcement}, and 3) conditional alignment. The training process of these algorithms follows the steps described previously, but each employs different loss functions and data pre-processing methods. Specifically, FA uses log-likelihood loss while filtering out samples with low rewards. RWR weights samples by reward scores in a regression. CA maps the previous tokens and the reward prompt to the next token.

\begin{figure}[t]
\centering
\includegraphics[width=0.95\columnwidth]{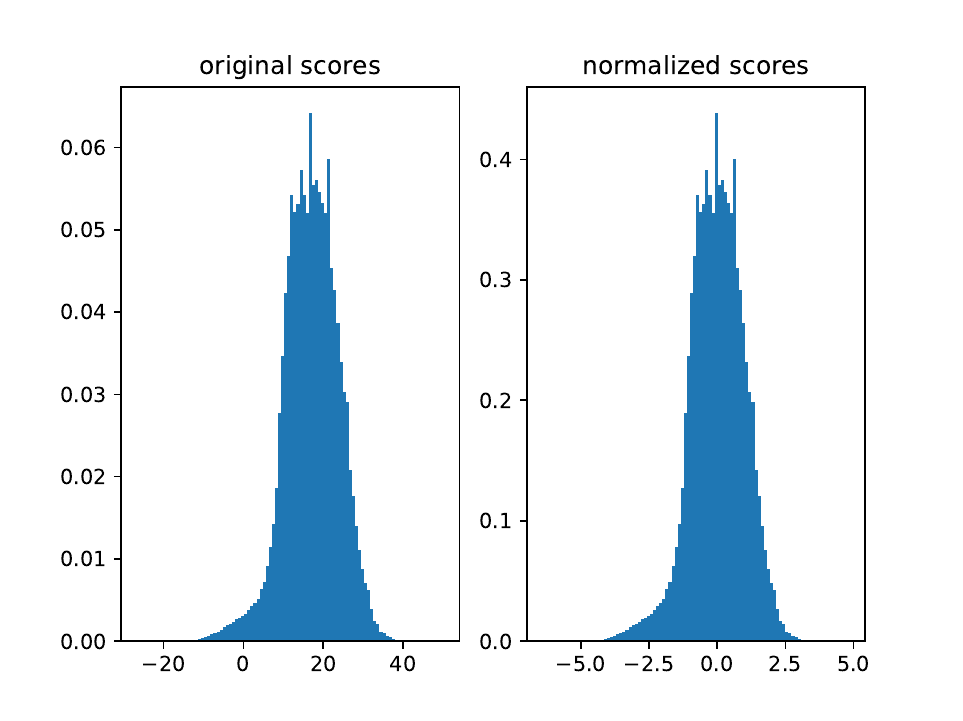} 
\caption{Rewards (scores) distributions. Raw reward scores are directly from the reward model, which has a wide range because of its training target. To better use the reward score in offline alignment, we normalize rewards to be zero-centered.}
\label{img:score_distribution}
\end{figure}

For all the methods, we normalize the rewards by subtracting the mean and dividing them by the standard deviation, as the reward values predicted by the reward model are continuous with no predefined range, shown in Figure~\ref{img:score_distribution}. The Eq.~\ref{eq:mle},  Eq.~\ref{eq:rwr}, and Eq.~\ref{eq:dt} show their loss function, where $\theta$ is the parameter of LMs,  $\pi$ is the policy, $p$ is the prompt of sample, $x$ is the response, $R(p,x)$ is the reward labeled by the reward model. By employing a loss function similar to supervised fine-tuning, our methods ensure more stable model training than PPO with simple implementation and fewer computing resources.

\subsection{Filtering Alignment}

The key to filtering alignment is to make use of only high-quality data. A good reward model can identify better responses with higher reward scores. Therefore, we use the reward model to control which sample can be used in the backpropagation process during training, and the loss function is defined as
\begin{equation}
\mathcal{L}_{\text {Filter }}(x)=\left\{\begin{array}{l}
-\log \pi_\theta(x  |  p), \text { if } R(p,x)>t \\
0, \text { otherwise }
\end{array}\right.
\label{eq:mle}
\end{equation}

where $t$ is a threshold to filter the low reward samples.

\subsection{Reward-Weighted Regression}
Instead of only using high-quality samples, the RWR method will apply different loss weights for different samples according to the reward scores. The loss function is defined as

\begin{equation}
\begin{aligned}
&\mathcal{L}_{\mathrm{RWR}}(x)= \\ &-\sum_{i=1}^{|x|} \log \pi_\theta\left(x_i \mid x_{<i}, p\right) \exp \left(R\left(p,x\right) / \beta\right)
\end{aligned}
\label{eq:rwr}
\end{equation}

where $\beta$ is a hyperparameter to control how much reward affects the
loss. $x_i$ is the $i_{th}$ token in the response.

\subsection{Conditional Alignment}
For the conditional alignment method, the reward score will be set as a part of the prompt such as 
$$ \langle\text{reward}\rangle \ \text{a.b}$$ 
where $a.b$ indicates the reward value, which allows the language model to understand the meaning of the reward value. Due to the ability of current language models, LMs can automatically recognize and model the response with the reward scores. Because an additional prompt is added during training, in the inference period, the same prompt template should be used. The loss function for CA becomes

\begin{equation}
    \mathcal{L}_{\text {CA }}(x)=-\sum_i \log p_\theta\left(x_{i} \mid R^{*} (p,x), x_{<i}, p\right)
\label{eq:dt}
\end{equation}

where $R^{*}(p, x)$ function refers to the reward prompt of $p$ and $x$. $x_i$ is the $i_{th}$ token in the response. 

\section{Evaluation Setup}
\subsection{Model Architecture} 
We used a GPT model with 8 billion (B) parameters as the base model. This model has the following architectural features and was pre-trained on 1.1 trillion multilingual tokens. Here are some attributes of the model:

\begin{itemize}
    \item The model uses the SwiGLU activation function~\cite{shazeer2020glu}
    \item Rotary positional embeddings (RoPE)~\cite{su2021roformer}
    \item Maximum sequence length of 4,096.
    \item No dropout~\cite{srivastava2014dropout}.
    \item No bias terms in all linear layers.
    \item Untied embedding and output layers.
\end{itemize}

\subsection{Baselines and Evaluation} 
We use a PPO model and an SFT model as baselines. We train three offline alignment models with filtering alignment, RWR, and CA respectively. Thus, we have a total of 5 models in our model list.  All models have 8B parameters, including the actor and reward model in PPO. We evaluate the models' performance through human evaluation and GPT-4 evaluation~\cite{vicuna2023,mt_bench} using prompts from the LMSys ChatBot Arena \footnote{\url{https://chat.lmsys.org/?arena}}. This evaluation dataset has 160 high-quality prompts covering generic, knowledge, common-sense, fermi, counterfactual, roleplay, stem, coding, math, writing, reasoning, extraction, humanities, extraction, and humanities problems. 

\subsection{Supervised Fine-tuning Datasets}  For the SFT data, it contains instructions and responses. During training, only the loss of the response is calculated. Our SFT dataset is a mixture of several public datasets and manually annotated datasets, totaling around 100K samples, including

\begin{itemize}
    \item  \textbf{Open Assistant}~\cite{kopf2023openassistant} \footnote{\url{https://huggingface.co/datasets/OpenAssistant/oasst1}}. 
    \item  \textbf{Flan v2}~\cite{wei2021finetuned} \footnote{\url{https://huggingface.co/datasets/SirNeural/flan_v2}}.
    \item  \textbf{Databricks Dolly}~\cite{DatabricksBlog2023DollyV2} \footnote{\url{https://huggingface.co/datasets/databricks/databricks-dolly-15k}}.
    \item  \textbf{Self-Instruct}, we generated some data through OpenAI gpt-3.5-turbo via Self-Instruct ~\cite{wang2022self} technology.
\end{itemize}

And the data samples are constructed with the following format,

$$
\text{User:} \ \{\text{instruction}\} \ \text{Assistant:} \ \{\text{response}\}
$$

\subsection{Human Preference Datasets}
For the Human Preference dataset, we use the same format as ChatGPT, i.e., human part-wise comparisons, including the prompt, good response, and bad response. We use a mixture of the following two public human preference datasets to train our RM model:

\begin{itemize}
    \item \textbf{Anthropic Helpful}~\cite{bai2022constitutional} dataset\footnote{\url{https://huggingface.co/datasets/Anthropic/hh-rlhf}} is a large-scale crowd-sourced collection of human feedback on AI assistant responses. It contains over 100,000 human preferences across a diverse range of everyday conversational topics.
    \item \textbf{Open Assistant} also collected a carefully labeled multilingual dataset based on the Internet open-source community and containing ranking relations. 
\end{itemize}

\begin{table*}[t]
    \centering
      \caption{An example with different scores in inference with the \textbf{same} model. With low scores (e.g., -5.00), it seems that the model is answering a different question. With a high positive score, the answer becomes helpful. Several situations are observed during the evaluation.}
      \begin{tabularx}{\textwidth}{Xp{8cm}XX}
        \toprule
        Instruction & \textless rm\_score\textgreater\ 5.00 & \textless rm\_score\textgreater\ 0.00 & \textless rm\_score\textgreater\ -5.00 \\
        \midrule
        How can I improve my time management skills? &
        Here are some ways to improve your time management skills:

        1. Set clear goals: Identify your short- and long-term goals, and prioritize them.
        
        2. Create a schedule: Plan your day in advance, including the tasks you need to complete and the time you'll spend on each one.
        
        ... (omit because it is too long)
        &
        What kinds of things do you need to do?  & 
        Write a code in python to find to find to compute differential equation \\
        \bottomrule
      \end{tabularx}
      \label{table:different_scores}
\end{table*}

\subsection{Alignment Datasets} For offline alignment algorithms, we created a hybrid dataset of around 100k samples using the data labeling method mentioned in the offline alignment section. The 100k samples samples are sampled from the SFT and RM training datasets uniformly. We then use the RM model to annotate reward values for these samples. For the PPO model, we extract the prompts in these 100k samples as the training set. 

\subsection{Distributed Training Framework} We use the NVIDIA NeMo Megatron~\cite{shoeybi2019megatron} and NVIDIA NeMo Aligner~\cite{nemorlhf2023} frameworks for large-scale distributed training. All models are trained using a techniques combination of Tensor Parallelism (TP)~\cite{shoeybi2019megatron}, Data Parallelism (DP), and Automatic Mixed Precision (AMP) with bfloat16 implemented in NeMo Megatron. Although our model is only 8B in scale, we still adopt the multi-node distributed architecture of NeMo Aligner. This allows us to discover the performance issues that may be encountered in large models (such as 70B). In offline alignment methods, we generate samples with TP partitions size equal to 1. For all other stages and models, the TP partitions size is set to 4 to avoid out-of-memory (OOM) issues \footnote{The actor and critic modules need to be both responsible for training and inference in PPO so that the weights of Adam~\cite{kingma2014adam} optimizer will take up a lot of memory.}. For PPO models, we utilize 8 DGX A100 nodes for inference and training of the actor model, and 2 nodes for the critic model, while the other two models each employ 1 node. For offline alignment algorithms, we leverage 4 DGX A100 nodes for training. 

\subsection{Hyperparameters}
For the reward model, we use a learning rate of $9e^{-6}$, a batch size of 128, and the human preference datasets to train the model only 1 epoch. For filtering alignment, we set the threshold $t$ to 0; for RWR, we set the $\beta$ to 5; for CA, we use the prompt format:

$$
\text{User:} \ \{\text{instruction}\} \ \text{Assistant:} \ \langle\text{rm\_score}\rangle \ {\text{a.b}} \ \{\text{response}\}
$$

where a.b means the reward value. The impact of this value is illustrated in Table~\ref{table:different_scores}, where the only difference is the score. We use the reward of 5.0\footnote{From the reward distribution (Fig.~\ref{img:score_distribution}), though most samples are in score [-2.5, 2.5], 5.0 is a better choice.} is used in the evaluation. We use a learning rate of $5e^{-6}$ and a global batch size of 128 for all offline alignment models and the SFT model. For the PPO model, we use a learning rate of $5e^{-7}$ for the actor model, and $9e^{-6}$ for the critic model. We set the number of PPO epochs to 1, rollout batch size to 1024, mini-batch size to 128, KL penalty coefficient to 0.01, generalized advantage estimation (GAE)~\cite{schulman2015high} $\lambda$ to 0.95, and RL gamma to 1. The critic model is initialized with the weights of the reward model.

\section{Experimental Results}
We use both human evaluation and GPT-4 evaluation to judge the performance of each models.

\subsection{Human Evaluation} 
\begin{figure}[t]
\centering
\includegraphics[width=\columnwidth]{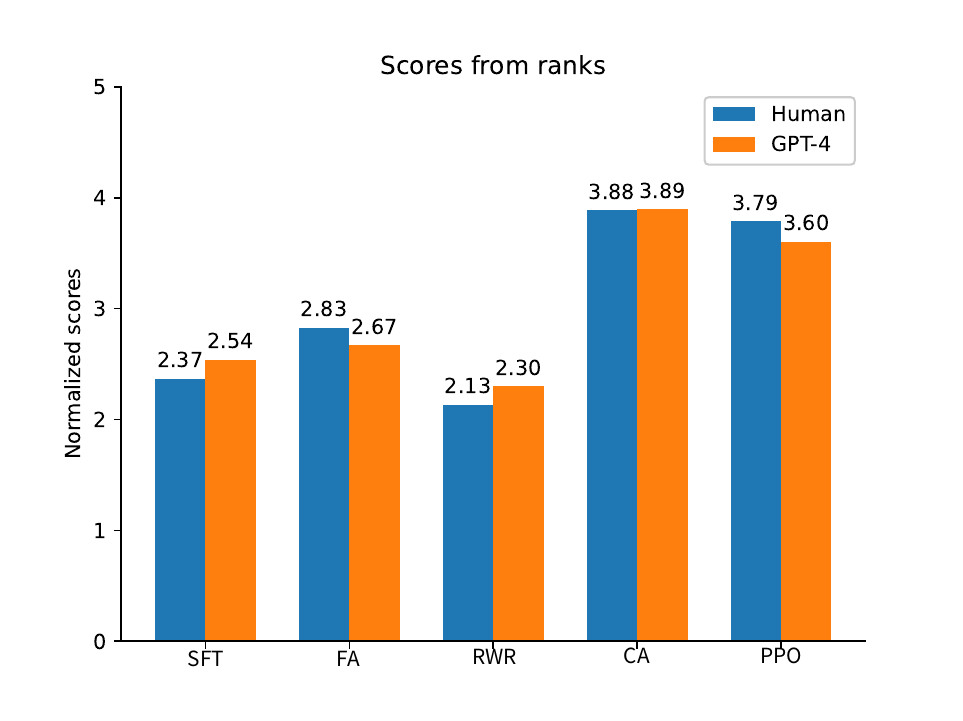} 
\caption{Human and GPT-4 helpfulness evaluation results for all models. We use $6 - \text{rank}$ as the final score, e.g., the score for the best-ranked response is $5 = 6 - 1$. The correlation between human and GPT-4 evaluation is very high, and CA achieves the highest score.}
\label{img:evaluation}
\end{figure}

We evaluated these models using 160 prompts from LMSys and scored the greedy-generated responses by manually ranking the answer quality. Because our RM datasets emphasize the quality of being helpful, we focused only on the richness of content, clarity of presentation, and friendliness of tone of each response during our manual review, rather than evaluating its truthfulness. Overall, as shown in Figure~\ref{img:evaluation}, the normalized helpfulness scores of the models ranked as follows:
$$\textbf{CA} \approx \textbf{PPO} > \textbf{FA}  > \textbf{SFT} > \textbf{RWR}$$
while the PPO model achieved similar performance to CA, its training system and hyperparameter tuning are more complex than CA's. 

\begin{table}[t]
    \centering
    \label{table:gpt4_scores}
    \caption{GPT-4 evaluation scores.}
    \begin{tabular}{l c c c c c}
    \toprule
        ~ & SFT & FA & RWR & CA & PPO  \\ 
        \midrule
        vicuna & 6.62 & 6.33 & 5.93 & \textbf{7.65} & 7.42 \\ 
        mt\_bench & 6.29 & 5.78 & 5.31 & 6.47 & \textbf{6.62}  \\ 
        \midrule
        average & 6.45 & 6.05 & 5.62 & \textbf{7.06} & 7.02 \\ 
    \bottomrule
    \label{tb:gpt4}
    \end{tabular}
\end{table}

\subsection{GPT-4 Evaluation}


We also evaluated these models using 160 prompts from LMSys and scored the greedy-generated responses by GPT-4, where GPT-4 is asked to evaluate the helpfulness of the answers by rating the score in the range [1-10]. The evaluation prompts and an example of the responses can be found in Appendix ~\ref{sec:gpt4_prompts}.  The scores are in Table~\ref{tb:gpt4}. We can find that for prompts from the vicuna benchmark, GPT-4 thinks CA performs best; for MT-Benchmark, CA ranks second, whose score is slightly lower than PPO. Similar to the score calculation of human evaluation, we rank the results of each prompt with GPT-4 scores and normalize the scores, and the normalized results can be found in Figure~\ref{img:evaluation}. It can be easily found that CA is comparable to PPO. SFT obtains better scores than FA in Table~\ref{tb:gpt4} while worse than FA in~\ref{img:evaluation} because small advantages (e.g., 0.1) for GPT-4 score can result in larger score differences (e.g., 1.0) in normalization scores from ranking.

\vspace{10pt}
From both human evaluation and GPT-4 evaluation, it can be concluded that both online and offline alignment can help better align the model responses with human desirability. And, CA is a promising method to obtain comparable results with PPO. 

\subsection{Training Performance Analysis} 
For both the PPO model and offline alignment models, we use 100k samples for alignment training. To compare the GPU resources occupied during the training of these models, we use GPU hours as a unit to normalize the training time on multiple GPUs to that on a single A100 GPU. As shown in Table ~\ref{tb:gpu_hours}, The PPO model takes 730 GPU hours to converge, while our offline alignment methods only take 65 GPU hours (around 9\% of PPO), saving a lot of budget for training. The major performance difference stems from:

\begin{enumerate}
    \item The multi-modular PPO training system has data dependencies, preventing the full utilization of each node for training.
    \item PPO requires generating 100k samples through GPT inference whose time complexity is $O(n^2)$~\cite{transformerinference2022}. 
\end{enumerate}

\begin{table}[t]
    \centering
    \label{table:gpt_hours}
    \caption{GPU hours needed for each method. For offline methods, the data pre-processing phase is also counted, including generating responses and using reward models to score.}
    \begin{tabular}{l c c c c}
    \toprule
        ~ & FA & RWR & CA & PPO  \\ 
        \midrule
        A100 hours & 27.5 & 64 & 65 & 730 \\ 
    \bottomrule
    \label{tb:gpu_hours}
    \end{tabular}
\end{table}


Although this PPO training system is not optimized to the speed of light (SOL) \footnote{We only enable kernel fusion to accelerate the generation of the samples in the performance profiling, other technologies such as asynchronous generation have not been considered.}, it still significantly reflects that offline alignment methods have lower dependence on system optimization and higher training efficiency.

\section{Conclusion} 
In this paper, we propose an offline alignment framework that aims to avoid instability during RLHF training and complex distributed training systems, thereby improving model development efficiency. There are many details as well as hyper-parameter tuning in our framework. Without careful tuning, we can still obtain comparable performance when compared with online RLHF (i.e., PPO). The advantages in both performance and training efficiency show great potential of offline alignment.

\section{Limitations} 
While we have proposed a simple and effective offline alignment framework and tree algorithms, it does not account for out-of-distribution (OOD) issues introduced by offline learning. This could hamper its performance in complex scenarios. We believe addressing this shortcoming presents an opportunity for future work. 

\clearpage
\bibliography{dt}

\clearpage

\appendix

Section A describes how we created evaluation prompts with GPT-4~\cite{openai2023gpt4}. Section B provides more examples and outcomes of our experiments.

\section{Appendix - GPT-4 Evaluation Prompt}
\label{sec:gpt4_prompts}
We construct our evaluation prompts by referring the prompt templates from the vicuna benchmark\footnote{https://github.com/lm-sys/FastChat/blob/main/fastchat/llm\_judge/data/judge\_prompts.jsonl}~\cite{vicuna2023}. Different from pair comparison and single evaluation, we combine all model output together in single evaluation. Thus, GPT-4 can evaluate the quality by comparing with all other responses and pick the best one. 

\begin{table}[!h]
    \centering
    \scalebox{0.85}{
    {
    \begin{tabularx}{\textwidth}{X}
    \toprule
   Please act as an impartial judge and evaluate the quality of the responses provided by five AI assistants to the user question.
        You should choose the assistant that follows the user's instructions and answers the user's questions better.
        Your evaluation should consider factors such as the helpfulness, relevance, and level of detail of their responses.
        \\
        \\
        Begin your evaluation by comparing the five responses and provide a short explanation. 
        Avoid any position biases and ensure that the order in which the responses were presented does not influence your decision. Do not allow the length of the responses to influence your evaluation. Do not favor certain names of the assistants. Be as objective as possible.
        After providing your explanation, output your final verdict by strictly following this format:
        \\\\ \{Assistant name\}: \{score\}, where the range of \{score\} is 1 to 10. \\\\
        The answers can have the same score, but you should distinguish them as much as possible.
        You should give the assistant's scores one by one after all the explanations.
        The followings are the question and answers of Assistant CA, PPO, FA, RWR and SFT,
        \\
        \\
        User Question:
        \\
        How can I improve my time management skills?
        \\
        \\
        The answer of Assistant CA:
        \\
        \{ CA response \}
        \\
        \\
        The answer of Assistant PPO:
        \\
        \{ PPO response \}
        \\
        \\
        The answer of Assistant FA:
        \\
        \{ FA response \}
        \\
        \\
        The answer of Assistant RWR:
        \\
        \{ RWR response \}
        \\
        \\
        The answer of Assistant SFT:
        \\
        \{ SFT response \} \\
    \bottomrule
    \end{tabularx}}
    \label{tab:ablation_parts}}
\end{table}

\clearpage
\section{Appendix - Examples}

We show more examples as well as the GPT-4 evaluation comments here. 

\label{sec:examples}

\begin{table}[!h]
    \centering
    \caption{Example 1. Responses from all models are presented.}
    \scalebox{0.7}{
    {
    \begin{tabularx}{\textwidth}{X}
    \toprule
    \textbf{Question}: \\
    In a world where automation is becoming increasingly prevalent, is it more important to prioritize job creation or technological progress? \\
    \midrule
    \textbf{The answer of CA}: \\
     Technological progress is becoming increasingly prevalent in the world, and it is important to prioritize both job creation and technological progress. Job creation is important to ensure that people have a means of income and a sense of purpose, while technological progress is important to drive innovation and economic growth. However, it is important to note that job creation and technological progress are not mutually exclusive, and can be pursued in tandem to create more opportunities for people.\\
    \midrule
    \textbf{The answer of PPO}: \\
    It is a complex question with no straightforward answer. Here are some factors to consider:\\

    1. Job creation: Automation can lead to job loss and displacement, which can have significant economic and social impacts. Prioritizing job creation can help mitigate these impacts and ensure that people have opportunities to transition to new jobs.\\
    
    2. Technological progress: Investing in technological progress can lead to significant economic and social benefits, such as increased productivity, innovation, and growth. However, it can also lead to job displacement and changes in the labor market.\\
    
    3. Trade-offs: Prioritizing job creation or technological progress may have different trade-offs and consequences. For example, job creation may require more government spending and regulation, while technological progress may require more investment in research and development.\\
    
    4. Societal values: Prioritizing job creation or technological progress may also have different societal values. Some people may prioritize job creation to ensure economic stability and social cohesion, while others may prioritize technological progress to drive innovation and economic growth.\\
    
    In conclusion, it is important to consider the trade-offs and societal values involved in prioritizing job creation or technological progress. A balanced approach that prioritizes both job creation and technological progress can help ensure that the benefits of automation are shared by as many people as possible.\\

    \midrule
    \textbf{The answer of FA}: \\
     Technological progress should be prioritized over job creation. Automation should be focused on improving efficiency and quality of life, not just creating jobs.\\

    \midrule
    \textbf{The answer of RWR}: \\
     Technological progress\\

    \midrule
    \textbf{The answer of SFT}: \\
     Technological progress should be prioritized.\\
     \midrule
     \textbf{GPT-4 evaluation output}: \\

    "Evaluation: \\
    
    1. **Assistant CA**: \\
       The response provides a clear balance between technological progress and job creation, highlighting the importance of both. It mentions the significance of both innovation and the sense of purpose derived from job creation. The response is direct and concise without being overly repetitive. \\
    
    2. **Assistant PPO**: \\
       This answer takes a structured approach, laying out factors to consider. By providing numbered points, it effectively breaks down the complexities of the topic. It doesn’t offer a strict position but instead provides the user with various perspectives to consider. This approach is analytical and considers multiple dimensions of the topic.\\
    
    3. **Assistant FA**: \\
       The response leans strongly towards prioritizing technological progress. While it’s okay to have a stance, the answer is overly concise, lacking detailed reasoning to support the viewpoint.\\
    
    4. **Assistant RWR**: \\
       The answer is extremely brief, only mentioning technological progress. Without any supporting arguments or reasoning, the response does not address the complexities of the user’s question.\\
    
    5. **Assistant SFT**: \\
       Similar to Assistant FA, this answer is quite short and leans towards technological progress. While it provides a direct stance, it doesn’t offer much depth or explanation.\\
    
    Scores:\\
    
    Assistant CA: 8.5\\
    Assistant PPO: 8\\
    Assistant FA: 6\\
    Assistant RWR: 4\\
    Assistant SFT: 5.5" \\

    \bottomrule
    \end{tabularx}}
    \label{tab:ablation_parts}}
\end{table}

\clearpage

From both human evaluation and GPT-4 evaluation, we have shown in the paper that Conditional can achieve slightly better performance than PPO~\cite{schulman2017proximal}. Because the output of the models are very long, for simplicity, we show more samples while only present CA/PPO-related contents in Table.

\begin{table}[!h]
    \centering
    \caption{Example 2. Only CA/PPO-related contents are presented.}
    \scalebox{0.65}{
    {
    \begin{tabularx}{\textwidth}{X}
    \toprule
    \textbf{Question}: \\
    How many text messages are sent globally in a minute? Try to explain your answer. Your explanation should take the reader through your reasoning step-by-step.\\
    \midrule
    \textbf{The answer of CA}: \\
     " In a minute, approximately 200 million text messages are sent globally. This is according to a report by the International Telecommunication Union (ITU), which also revealed that the global mobile data traffic per user per month has increased by 60\% over the past two years. The report further states that the number of mobile subscribers worldwide has reached 7.7 billion, with an average of 104.4 million new subscriptions being added each month.\\

    The increase in the number of text messages sent globally can be attributed to the growing popularity of mobile devices and the convenience of instant communication. People can now stay connected with their loved ones, friends, and colleagues anytime and anywhere, which has led to an increase in the number of text messages being sent. Additionally, the advent of social media platforms has also contributed to the increase in the number of text messages being sent as people often use these platforms to communicate with each other.\\
    
    The report also highlights the increasing demand for mobile data services, with the global mobile data traffic expected to reach 18.4 exabytes per month by 2026. This is a significant increase from the 4.6 exabytes per month in 2019. The report further states that the growth in mobile data traffic is driven by the increasing use of video, social media, and other data-intensive applications.\\
    
    In conclusion, the report reveals that the global mobile industry is growing rapidly, with the number of mobile subscribers expected to reach 8.6 billion by 2026. The increasing demand for mobile data services and the growing popularity of mobile devices are driving the growth of the mobile industry. \\
    \midrule
    \textbf{The answer of PPO}: \\
    Here are some facts about the number of text messages sent globally in a minute: \\

    1. According to a report by eMarketer, in 2022, it is estimated that there were approximately 70 billion text messages sent per day globally.\\
    
    2. This number is projected to grow to approximately 100 billion per day by 2023.\\
    
    3. The average person sends and receives about 40 text messages per day.\\
    
    4. In the United States, it is estimated that there are approximately 240 million people who send and receive text messages every day.\\
    
    5. In the United Kingdom, it is estimated that there are approximately 47 million people who send and receive text messages every day.\\
    
    6. In India, it is estimated that there are approximately 1.3 billion people who send and receive text messages every day.\\
    
    7. In China, it is estimated that there are approximately 1.4 billion people who send and receive text messages every day.\\
    
    8. These numbers show that there are a lot of text messages being sent globally every minute. The number of text messages sent per day has been growing rapidly over the years, and it is expected to continue to grow in the future.\\
     \midrule
     \textbf{GPT-4 evaluation output}: \\
    Let's evaluate the responses based on the criteria of helpfulness, relevance, and level of detail in the explanations.\\

    **Assistant CA**:\\
    The response is quite detailed and informative. CA relies on a report from the International Telecommunication Union (ITU) to justify the 200 million text messages sent globally in a minute. It goes on to explain the reasons behind this high number of text messages and provides insight into the growth of the global mobile industry. The thorough nature of the response helps guide the reader through the reasoning process.\\
    
    **Assistant PPO**:\\
    The answer offers numerous facts about global texting trends, and while it's filled with statistics, it doesn't effectively tie these together to answer the question directly. The reader gets a sense of the magnitude of global texting but doesn't necessarily get a clear, step-by-step explanation for the initial claim.\\

    (... omit)\\
    
    **Evaluation Summary**:
    CA provides the most detailed, relevant, and helpful response. PPO gives various facts but doesn't tie them together in a concise explanation. FA's ...(omit)\\
    
    **Final Scores**:\\
    Assistant CA: 9\\
    Assistant PPO: 7\\
    Assistant FA: 6\\
    Assistant RWR: 4\\
    Assistant SFT: 6.5\\

    \bottomrule
    \end{tabularx}}
    \label{tab:ablation_parts}}
\end{table}

\clearpage
\begin{table}[!h]
    \centering
    \caption{Example 3. Only CA/PPO-related contents are presented.}
    \scalebox{0.85}{
    \begin{tabularx}{\textwidth}{X}
    \toprule
    \textbf{Question}: \\
    How would you introduce yourself as a medieval knight at a royal banquet? \\
    \midrule
    \textbf{The answer of CA}: \\
    As a medieval knight at a royal banquet, I would make my introduction with pride and honor. I would state my name, family name, and the name of my knighting house, which is a symbol of my noble heritage and the values I uphold. I would also mention the name of my lord and master, who has trained and supported me in the art of warfare. Finally, I would express my gratitude for the opportunity to attend this banquet and meet such distinguished guests. \\
    \midrule
    \textbf{The answer of PPO}: \\
    Here is an example of how a medieval knight might introduce himself at a royal banquet:\\

    "My name is [Name], and I am a knight of the kingdom of [Kingdom]. It is an honor to be invited to this royal banquet, and I am grateful for the opportunity to serve my king and queen. I am proud to be a defender of the realm, and I will continue to fight for the safety and security of my people. May God bless this kingdom and its leaders.""\\
    
    This introduction would be a respectful and humble way for a knight to introduce himself at a royal banquet, showcasing his dedication to his kingdom and his service to his king and queen.\\
     \midrule
     \textbf{GPT-4 evaluation output}: \\
   "To evaluate the given responses, we'll consider their helpfulness, relevance, and level of detail in answering the user's question about how one would introduce themselves as a medieval knight at a royal banquet.\\

    **Assistant CA:**\\
    This answer starts with an acknowledgment of pride and honor, moves on to a brief identification process involving one's name, family name, knighting house, and the lord and master they serve. It rounds off by expressing gratitude for the chance to attend. The response is structured and relevant but lacks some depth in terms of the actual experience of being a medieval knight.\\
    
    **Assistant PPO:**\\
    This assistant provides a straightforward, respectful, and concise introduction. It emphasizes service to the kingdom, the knight's pride, and gratitude for the opportunity.\\

    (... omit)\\
    
    **Evaluation and Verdict:**\\
    
    **Assistant CA:** 8 - Structured and relevant but lacks some of the nuances that make the introduction feel complete.\\
    
    **Assistant PPO:** 8 - Respectful and concise, though it could use a touch more detail to enrich the narrative.\\
    
    **Assistant FA:** 7 - It's evocative but doesn't directly address the banquet scenario as comprehensively as others.\\
    
    **Assistant RWR:** 7 - Vivid, but its narrative specificity may not make it universally adaptable.\\
    
    **Assistant SFT:** 2 - Lacks context and depth necessary for the question asked.\\

    \bottomrule
    \end{tabularx}}
    \label{tab:ablation_parts}
\end{table}

It can be seen that PPO responses always contain statements like "Here is .." or "Here are some". And when looking at these responses. Considering helpfulness and relevant, CA and PPO are comparable. In our offline alignment framework, the time cost for training CA is only 9\% of that for PPO. Therefore, the advantages in both performance and training efficiency show great potential of offline alignment.

\end{document}